\documentclass{article}

    \PassOptionsToPackage{numbers, compress}{natbib}

\usepackage[preprint]{neurips_2025}

\usepackage{booktabs}
\usepackage{algorithm}
\usepackage{algorithmic}
\usepackage{subcaption}
\usepackage{graphicx}
\usepackage{amsmath}
\usepackage{makecell}
\usepackage{enumitem}
\usepackage{hyperref}
\usepackage{wrapfig}

\usepackage{caption}
\usepackage{booktabs}

\usepackage[capitalize]{cleveref}
\crefname{section}{Sec.}{Secs.}
\Crefname{section}{Section}{Sections}
\Crefname{table}{Table}{Tables}
\crefname{table}{Tab.}{Tabs.}

\title{Code-as-Room: Generating 3D Rooms from Top-Down View Images via Agentic Code Synthesis}

\author{
    Yixuan Yang$^{1}$\footnotemark[1]\quad
    Zhen Luo$^{2, 3}$\footnotemark[1]\quad
    Wanshui Gan$^{1}$\footnotemark[1]\quad
    Jinkun Hao$^{1}$\quad \\
    \textbf{Junru Lu$^{4}$\quad
    Jinghao Yan$^{1}$\quad
    Zhaoyang Lyu$^{1}$\quad
    Xudong Xu$^{1}$\footnotemark[2]\quad}
    \\[1mm]
    $^{1}$ Shanghai Artificial Intelligence Laboratory \quad
    $^{2}$ Shanghai Innovation Institute\\
    $^{3}$ Southern University of Science and Technology \quad
    $^{4}$ University of Warwick\\
    \small Project page: \url{https://code-as-room.github.io/}
}

\begin{document}
\footnotetext[1]{Equal contribution.}
\footnotetext[2]{Corresponding author.}
\footnotetext[3]{Contact: \texttt{arnoldyang97@gmail.com}.}
\maketitle

\begin{figure*}[h]
    \vspace{-2em}
    \centering
    \includegraphics[width=1\linewidth]{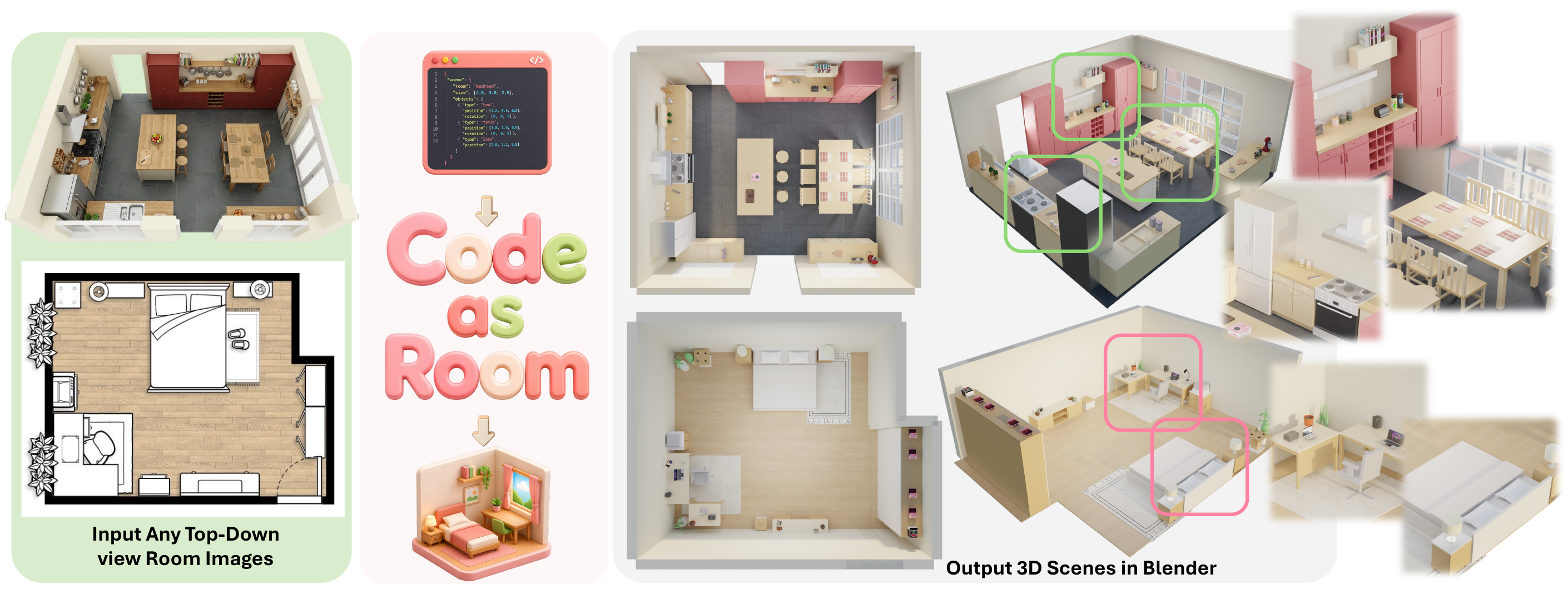}
    \caption{\textbf{Code-as-Room} brings diverse interactive 3D scenes from a single top-down view image. We design an agentic system with a structured execution harness and activate the MLLMs' ability to understand, design, and code the 3D rooms in Blender. }
    \label{fig:teaser}
    \vspace{-0.6em}
\end{figure*}

\begin{abstract}
Designing realistic and functional 3D indoor rooms is essential for a wide range of applications, including interior design, virtual reality, gaming, and embodied AI.
While recent MLLM-based approaches have shown great potential for 3D room synthesis from textual descriptions or reference images,
text-based methods struggle to capture precise spatial information, and existing image-conditioned agents suffer from instability and infinite looping when tasked with holistic room generation from top-down views.
To address these limitations, we propose \textbf{Code-as-Room}, an MLLM-based agentic framework equipped with a structured execution harness, which represents 3D rooms with Blender codes.
Given a top-down room image, the framework parses the reference image to extract scene elements and their spatial relationships, and synthesizes executable Blender code for geometry, materials, and lighting in a principled, multi-stage pipeline.
A cross-stage memory module is maintained throughout to mitigate context forgetting inherent to existing agent-based frameworks.
We further introduce a dedicated benchmark for code-based 3D room synthesis, encompassing various evaluation protocols.
Based on our benchmark, comprehensive comparisons against existing agent-based methods are conducted to validate the effectiveness of our proposed execution harness.
\end{abstract}

\vspace{-1em}
\section{Introduction}
\label{sec:intro}

Designing realistic and functional 3D indoor rooms plays a crucial role in the interior design, virtual reality, games, and even embodied AI~\cite{armeni20163d,silberman2012indoor,li2020openrooms,li2018interiornet}.
However, manually constructing 3D rooms is labor-intensive, requiring expertise in 3D object modeling, spatial arrangement, material design, and light adjustment.
Traditional graphics methods have explored procedural generation, rule-based layout optimization, and constraint-driven object placement to reduce manual effort in room creation~\cite{xu2002constraint,ProcTHOR,balint2019generalized,Makeithome, merrell2011interactive,yu2015clutterpalette}.
Admittedly, these methods are severely limited by hand-crafted rules and predefined categories, failing to handle complex real-world spatial relationships and flexible user needs.

Recently, multimodal large language models (MLLMs)~\cite{openai2026gpt55,gemini31pro2026} have experienced significant prosperity, demonstrating remarkable performance and strong generalization across various application scenarios. Thus, leveraging the power of MLLMs for 3D room generation has become an intriguing research problem.
A line of scene generation approaches~\cite{feng2023layoutgpt,yang2024holodeck,ccelen2024design,fu2024anyhome,yang2024llplace,sun2025layoutvlm,yang2026optiscene,luo2026stable} represents indoor scenes as JavaScript Object Notation (JSON), or other structured formats, and resorts to MLLMs to predict spatial layout, inter-object relations, and object attributes according to users' scene descriptions.
Afterwards, 3D objects are retrieved or generated based on their attributes to compose a complete 3D room.
Nevertheless, text descriptions of scenes fall short of specifying interior spatial information, such as object counts, precise locations, or detailed orientations, resulting in generated 3D rooms that fail to align with users' preferences.

In real-world design workflows, top-down layouts, sketches, and floor-plan-like images are widely employed to facilitate 3D room creation, since they inherently encode rich spatial priors and holistic scene appearance.
In practice, 3D designers iteratively refine their work by consulting these references until a satisfactory result is achieved.
Following this paradigm, researchers~\cite{yang2026sceneweaver,viga,xia2026sage,pfaff2026scenesmith, 3DGeneralist} have explored leveraging MLLM agents to synthesize 3D rooms from reference images, where the agent alternates between a generator role and a critic role in an iterative manner.
Notably, a seminal work, VIGA~\cite{viga}, proposes an intriguing coding agent that generates executable Blender code to construct 3D scenes, encompassing both the scene layout and the constituent 3D objects.
Given a perspective image as input, VIGA reconstructs the corresponding 3D scene via an analysis-by-synthesis loop, demonstrating promising potential for the code-as-scene paradigm.
However, when naively extending from perspective images to top-down views for complete 3D room generation, VIGA struggles to recover fine-grained spatial details.
More critically, the agent is prone to falling into infinite loops, resulting in unstable and unreliable generation outcomes.

To circumvent this hurdle, we propose \textbf{Code-as-Room} (CaR), an MLLM-based agentic framework equipped with \emph{a structured execution harness} for top-down image-based 3D room synthesis, which generates executable Blender code to represent 3D rooms.
The framework first parses the top-down reference image to identify three categories of scene elements: major furniture, small accessory items attached to them, and interior finishes comprising doors, windows, and walls. 
Their spatial interrelations are inferred from the image to form a holistic scene graph representation.
The framework then generates layout code for all identified elements and iteratively refines the overall spatial arrangement through a render-and-compare feedback loop, empowered by the visual recognition capabilities of the MLLM. 
Building upon the refined layout, the MLLM is further employed to profile each 3D object by inferring detailed properties including appearance, functional attributes, and material characteristics. Following object profiling, the framework enters the object code generation phase, in which the agent synthesizes Blender code for both geometry and surface materials, strictly conditioned on the inferred object profiles.
Asset retrieval and 3D object generation are incorporated as auxiliary modules to handle challenging small items with complex geometric details.
Finally, texture, material, and lighting codes are generated to accomplish interior decoration, substantially enhancing the perceptual aesthetics and photorealistic fidelity of the synthesized room.
Notably, a cross-stage memory module is maintained throughout the entire pipeline to store the outputs of each stage, effectively mitigating the pervasive context forgetting problem inherent to existing agent-based frameworks.

To systematically evaluate existing MLLM models and our agentic framework, we construct a dedicated benchmark for code-based 3D Room synthesis.
Beyond assessing visual quality, the benchmark is designed to evaluate the distinctive challenges of this task, encompassing visual content understanding, spatial relationship reasoning, and vision-to-code generation capability. 
Moreover, comprehensive comparisons against existing agent-based methods validate the effectiveness of our proposed harness for code-based 3D room generation.
Overall, our contributions can be summarized as follows:
\begin{itemize}[topsep=2pt, itemsep=2pt, parsep=0pt]
    \item We propose a top-down image-guided paradigm for 3D room synthesis, where the input image serves as a global spatial prior to guide complete indoor room generation.
    \item We propose a structured execution harness that orchestrates the MLLM agent for code-based 3D room generation, ensuring stable and coherent 3D room synthesis.
    \item We introduce an Image-to-3D Room synthesis benchmark and conduct comprehensive experiments to evaluate the existing MLLMs in terms of visual understanding, spatial reasoning, vision-to-code ability, and scene quality.
\end{itemize}

\section{Related Work}
\label{sec:related}
\paragraph{Procedural and Data-driven Indoor Scene Synthesis}
Indoor scene synthesis has long been studied in computer graphics. Early methods typically formulate scene generation as a rule-based, constraint-based~\cite{merrell2011interactive}, or optimization-driven problem~\cite{Makeithome}. For example, constraint-based placement systems allow users to compose complex scenes by specifying semantic and geometric constraints, while furniture layout methods incorporate interior design guidelines, ergonomic objectives, and spatial priors to optimize plausible object arrangements. Other works learn arrangement priors from example scenes, synthesizing new 3D object configurations by modeling object co-occurrence, support relations, and spatial distributions. Beyond major furniture layout, interactive tools such as ClutterPalette~\cite{yu2015clutterpalette} further support the placement of small-scale objects to enrich indoor scenes. More recently, procedural generation frameworks such as ProcTHOR~\cite{ProcTHOR} have scaled indoor environment construction to large numbers of interactive houses for embodied AI training and evaluation.
However, these methods are mostly driven by rules, constraints, examples, or procedural templates, leaving the problem of generating a complete, editable, and executable 3D room from a room-level visual input largely unexplored.

\paragraph{LLM- and Agent-based 3D Scene Generation}
Large language models have recently been used for 3D scene generation due to their commonsense reasoning and open-vocabulary planning ability. Methods such as Holodeck~\cite{yang2024holodeck}, LLplace~\cite{yang2024llplace}, LAYOUTVLM~\cite{sun2025layoutvlm} and I-Design~\cite{ccelen2024design} generate room layouts, object selections, spatial relations, or scene graphs from language instructions, demonstrating the potential of LLMs for controllable indoor scene synthesis. Building on this direction, recent agentic frameworks introduce tool use, feedback loops, and multi-agent collaboration. For example, SceneWeaver~\cite{yang2026sceneweaver} employs a self-reflective agent to coordinate different generation tools and iteratively refine scenes, SAGE~\cite{xia2026sage} couples scene generators with critics to produce simulation-ready environments for embodied AI, and SceneSmith~\cite{pfaff2026scenesmith} uses hierarchical VLM agents to progressively construct indoor scenes from architectural layout to furniture and small objects.
Despite these advances, existing LLM- and agent-based methods are still primarily text- or task-driven. In practical design scenarios, however, users often start from floor plans, top-down sketches, or layout images, where room structure and object arrangements are already visually specified. This makes room-level image input a more direct and valuable condition for 3D scene generation, yet it remains underexplored in existing agentic frameworks.

\paragraph{Image-conditioned 3D Generation and Code-based 3D Scene Representation}
Image-conditioned 3D generation has recently achieved impressive progress, with many methods generating 3D assets from images using diffusion models, neural fields, meshes, or other learned representations~\cite{liu2023zero, liu2024syncdreamer,yao2025cast}. However, most of these methods mainly focus on single objects or relatively simple scenes, and their outputs are often designed for reconstruction or visual synthesis rather than the structured generation of a complete indoor room. As a result, they are not well-suited for modeling room-scale scene elements such as global layout, major furniture, minor objects, material appearance, and lighting in a unified way.
A particularly relevant work is VIGA~\cite{viga}, which demonstrates the potential of using code to represent 3D structure from visual input. However, its generation pipeline with inadequate harness design and does not address the synthesis of an entire room with complex vision input.
More broadly, recent code-based 3D generation methods~\cite{3d-gpt, jones2020shapeassembly,lu2025ll3m} have shown that executable programs are a promising representation for 3D content due to their interpretability and editability. Nevertheless, these approaches mainly focus on individual objects or localized structures, rather than representing a complete room-scale scene in code.
In contrast, our work aims to generate a complete indoor room from a top-down image, where the full scene is represented as executable Blender code.

\begin{figure*}[t]
  \centering
    \includegraphics[width=1\textwidth]{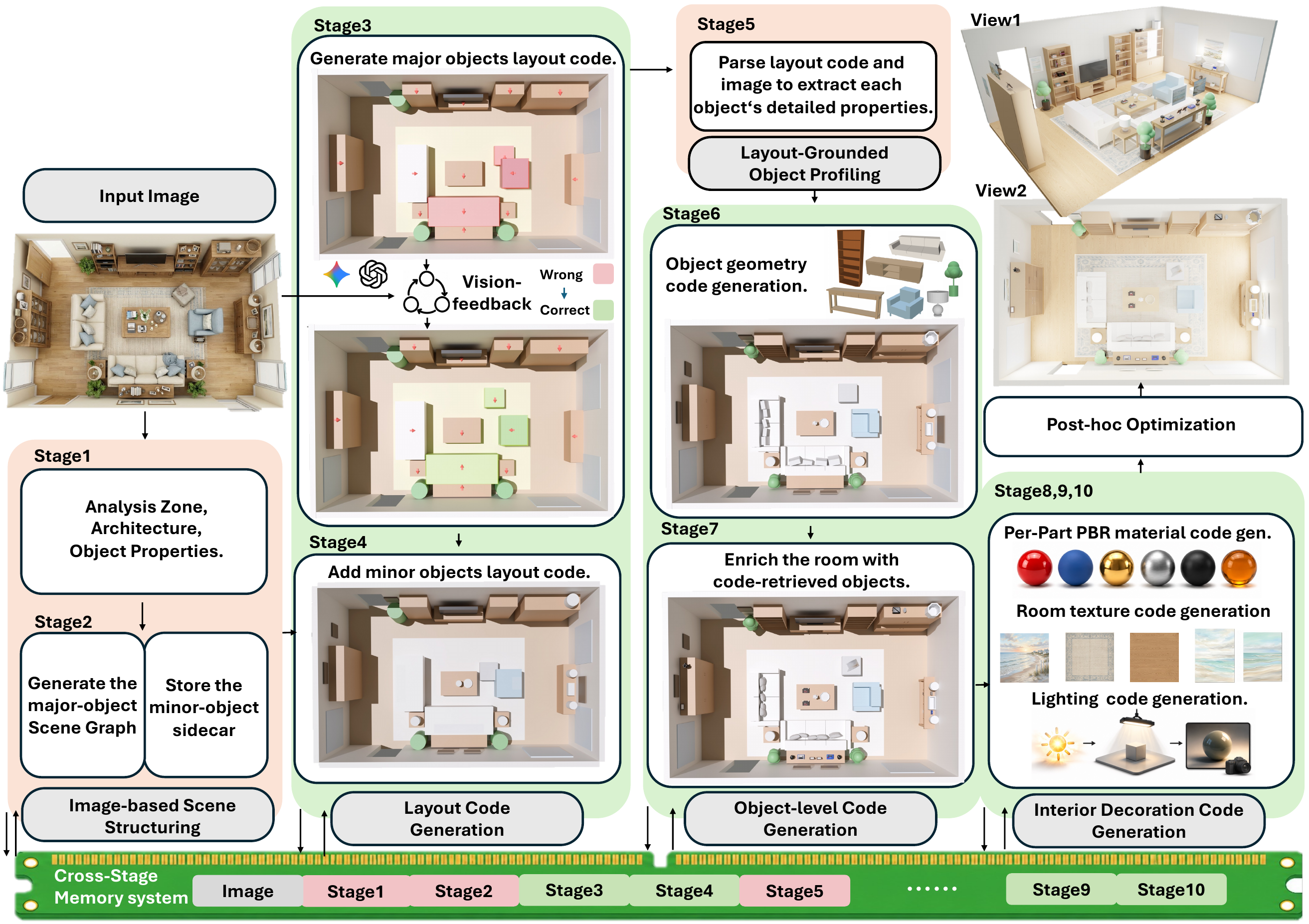}
  \caption{
    Overview of the \textbf{Code-as-Room} pipeline. A single top-down view image is progressively transformed into a fully renderable 3D scene through a sequence of specialized MLLM agent stages, organized into five phases: image-based scene structuring, layout code generation, layout-grounded object profiling, object-level code generation, and interior decoration code generation. Arrows denote data flow through the cross-stage memory system, wherein each stage reads upstream outputs and writes its own results as typed memory entries.
  }
  \label{fig:pipeline}

\end{figure*}
\section{Method}
\label{sec:method}

\subsection{Problem Definition}

Given a room-level top-down image $I$, our goal is to generate executable Blender code $C$ that constructs a complete 3D indoor scene aligned with the input image. The code specifies room structure, object placement, object geometry, materials, lighting.
We formulate the task as an agentic image-to-code generation process:
\[
C = \mathcal{A}(I),
\]
where $\mathcal{A}$ denotes the proposed VLM-agent harness.

Directly generating complete scene code from a single image is challenging because the model must jointly infer room structure, spatial layout, object geometry, and appearance. We therefore decompose the process into a coarse-to-fine workflow. The coarse stage builds a structured scene state and a layout program:
\[
D_{\mathrm{CU}} = U_{\mathrm{CU}}(I,\mathcal{M}), \qquad
C_{\mathrm{layout}} = G_{\mathrm{CG}}(I,D_{\mathrm{CU}},\mathcal{M}),
\]
where $D_{\mathrm{CU}}$ denotes the coarse scene understanding result and $C_{\mathrm{layout}}$ is the executable layout code. The fine stage then enriches the layout with image-grounded object descriptions and synthesizes the final room program:
\[
D_{\mathrm{FU}} = U_{\mathrm{FU}}(I,C_{\mathrm{layout}},\mathcal{M}), \qquad
C = G_{\mathrm{FG}}(I,C_{\mathrm{layout}},D_{\mathrm{FU}},\mathcal{M}).
\]
Here, $\mathcal{M}$ is the cross-stage memory shared by all modules. This decomposition separates global spatial alignment from local detail synthesis: the coarse stage fixes room structure and object placement, while the fine stage adds editable geometry, appearance, and rendering code.
The whole pipeline of Code-as-Room is shown in Fig.~\ref{fig:pipeline}.

\subsection{Cross-Stage Memory}
\label{sec:memory}

We maintain a shared memory $\mathcal{M}$ as the persistent state of the pipeline. 
After stage $s$ produces a typed artifact $e_s=\langle s,\tau_s,O_s,\eta_s\rangle$, 
memory is updated by
\[
\mathcal{M}_s=\mathcal{M}_{s-1}\oplus e_s .
\]
Each downstream stage reads only a predefined memory view, preserving cross-stage consistency while reducing prompt noise and hallucinated dependencies.

\subsection{Image-based Scene Structuring}
\label{sec:coarse-understanding}

The first two stages convert the top-down image $I$ into a structured scene state for layout generation. Stage 1 extracts a schema-constrained description $D_1$, and Stage 2 builds an object-centric scene graph $G=(V,E)$ with a minor-object sidecar $M_{\mathrm{minor}}$.

\paragraph{Stage 1: Spatial semantic analysis.}
The VLM outputs $D_1$ with functional zones, object hierarchies, and architectural elements. Each object is assigned an identifier, category, placement type, and parent when applicable, while walls, doors, windows, openings, and built-in structures are kept as fixed spatial references. A perimeter-aware prompt scans walls, corners, openings, and a coarse grid to recover peripheral and wall-mounted objects. The resulting description is
\[
D_1 = F_1(I,P_1), \qquad
\mathcal{M} = \mathcal{M} \oplus D_1 .
\]

\paragraph{Stage 2: Object-centric scene graph construction.}
Stage 2 reads $\{I,D_1\}$ and first derives a deterministic skeleton
\[
S=\{V_{\mathrm{arch}},V_{\mathrm{major}},E_{\mathrm{parent}},M_{\mathrm{minor}}\},
\]
where $V_{\mathrm{arch}}$ contains architectural features, $V_{\mathrm{major}}$ contains layout-defining objects, $E_{\mathrm{parent}}$ contains hierarchy-derived relations, and $M_{\mathrm{minor}}$ stores minor objects for later placement. The VLM only completes attributes, geometry hints, and forward relations among existing nodes. After filtering invalid edges and adding inverse and architectural-anchor relations, the graph and sidecar are written to memory:
\[
E=E_{\mathrm{parent}}\cup E_{\mathrm{vlm}}\cup E_{\mathrm{wall}}\cup E_{\mathrm{corner}}\cup \mathrm{Inv}(\cdot),
\quad
\mathcal{M}=\mathcal{M}\oplus\{G,M_{\mathrm{minor}}\}.
\]

\subsection{Layout Code Generation}
\label{sec:coarse-layout}

Given $D_1$, $G=(V,E)$, $M_{\mathrm{minor}}$, and $I$, this stage generates a coarse layout program $C_{\mathrm{layout}}$. Objects are instantiated as named bounding-box proxies in Blender with approximate position, scale, and orientation, while detailed geometry, materials, lighting, and tiny objects are deferred. Because each placement is emitted as a primitive-constructor call, $C_{\mathrm{layout}}$ can be rendered for feedback and parsed by later stages.

We use two sub-stages. The major sub-stage generates the floor-level arrangement of layout-defining furniture from $D_1$ and $G$. The auxiliary sub-stage freezes the major layout and appends wall-mounted objects $D^{\mathrm{wall}}_1=\{o\in D_1\mid o.\mathrm{placement\_type}=\mathrm{wall}\}$ and visually salient minor objects from $M_{\mathrm{minor}}$.

\paragraph{Stage 3: Major layout with visual feedback.}
We refine the major layout through a render--critique--revise loop initialized by
\[
C^{(0)}=\mathrm{Generate}(I,D_1,G).
\]
At each iteration, the current code is rendered into a top-down image and evaluated by a VLM-based critic:
\[
\begin{aligned}
R^{(t)} &= \mathrm{Render}(C^{(t-1)}),\\
(A^{(t)},s_t) &= \mathrm{Critique}(I,R^{(t)},G),\\
\widetilde{A}^{(t)} &= \mathrm{Sanitize}(A^{(t)},D_1,G),\\
C^{(t)} &= \mathrm{Revise}(C^{(t-1)},\widetilde{A}^{(t)}).
\end{aligned}
\]
Here, $s_t$ denotes the VLM-assessed layout quality score, which summarizes object coverage, overlap, boundary consistency, and spatial relation correctness based on the rendered view.
The critic also outputs textual feedback $A^{(t)}$ describing missing objects, overlaps, boundary violations, and relation errors.
Since the critic may occasionally suggest unsupported architectural changes, we sanitize its feedback with respect to $D_1$ and $G$ before revising the code.
The loop terminates once $s_t\ge s^\star$ or the maximum number of iterations $T_{\max}$ is reached, where we set $T_{\max}=5$ in our experiments.
The final output is denoted as $C_{\mathrm{layout}}^{\mathrm{major}}$.

\paragraph{Stage 4: Auxiliary layout for walls and salient minors.}
Stage~4 complements the major layout with wall-mounted objects and visually salient minor objects.
Wall-mounted objects are aligned to the inferred wall planes according to their semantic anchors in the scene graph.
For minor objects, we only keep items that are visible and layout-relevant at the coarse scene scale:
\[
M^{\star}_{\mathrm{minor}}
=
\{m \in M_{\mathrm{minor}}\},
\]
where $m$ is visible at the coarse layout scale and not surface-bound.
These objects include rugs, floor lamps, plants, and large decorations.
Tiny surface-bound objects, such as books, cups, and small tabletop items, are deferred to the later fine-grained placement stage, where they are placed according to supporting surfaces and the memory from Stage~2.
The selected wall and minor objects are appended as primitive-constructor calls:
\[
C_{\mathrm{layout}}
=
\mathrm{Append}
\left(
C^{\mathrm{major}}_{\mathrm{layout}},
M^{\star}
\right),
\qquad
\mathcal{M}\leftarrow \mathcal{M}\oplus C_{\mathrm{layout}} .
\]
The resulting layout serves as the scaffold for fine-grained description, geometry generation, and small-object placement.

\subsection{Object-level Code Generation}
\label{sec:stage8}

After layout code is fixed, the object-level module enriches each proxy with image-grounded appearance, procedural part geometry, and surface-bound small objects. 

\paragraph{Stage 5: Layout-grounded object description.}
The coarse layout fixes each object’s identifier, category, pose, and size, but lacks visual details for geometry, materials, and textures. We parse $C_{\mathrm{layout}}$ into placed objects and use their layout attributes to ground the VLM. Conditioned on the input image and memory, the VLM produces fine-grained object descriptions $D_{\mathrm{FU}}$ covering color, material, function, structure, and style, together with a global room-style description JSON file $s_{\mathrm{room}}$. These outputs preserve fixed placement and are written to memory.

\paragraph{Stage 6: Object geometry replacement.}
For each placed object $o_i$ in $C_{\mathrm{layout}}$, the geometry agent predicts a semantic 3D geometry primitive decomposition:
\[
\mathcal{P}_i=\Phi_{\mathrm{geo}}(o_i,d_i)=\{p_{i,j}\}_{j=1}^{K_i},
\]
where $d_i\in D_{\mathrm{FU}}$, and each part $p_{i,j}$ specifies a 3D geometry primitive type, semantic part name, local size, offset, and rotation. Since parts are defined in the local frame of the original proxy, the generated object inherits the coarse-layout pose. We replace proxy constructors in the layout program with part-based constructors:
\[
C_{\mathrm{geom}}
=
\mathrm{Replace}
\left(
C_{\mathrm{layout}},
\{o_i\mapsto \mathcal{P}_i\}_{i=1}^{N}
\right).
\]
The same mapping is stored as a geometry dictionary for later surface discovery and material assignment.

Tiny objects are instantiated through a hybrid procedural-and-retrieval strategy.
For visually distinctive objects, we first create simple geometric placeholders to preserve their positions and occupied regions, and then retrieve matching assets from an asset library $\mathcal{B}$ to replace these placeholders.
The selected asset is obtained by
\[
b^\star
=
\arg\max_{b\in\mathcal{B}}
\mathrm{match}
\left(
b;\text{label},\text{description},\text{placeholder size}
\right),
\]
where the matching score jointly considers semantic relevance and size compatibility.
The selected asset is scaled and aligned to the corresponding placeholder, preserving its support surface and footprint.
And the surface discovery and occupancy detection algorithm can be found in the Appendix.

\subsection{Interior Decoration Code Generation}
\label{sec:stage85}

After object-level generation, we complete appearance and illumination through geometry-preserving code rewriting: 
\[
C_{\mathrm{obj}}
\xrightarrow{\mathrm{ApplyMat}}
C_{\mathrm{mat}}
\xrightarrow{\mathrm{ApplyTex}}
C_{\mathrm{tex}}
\xrightarrow{\mathrm{RenderSetup}}
C_{\mathrm{raw}},
\]
where each step rewrites or appends Blender code without modifying object placement or geometry.

\paragraph{Stage 8: Material assignment.}
Since objects are decomposed into semantic parts, we assign part-level PBR materials using the part dictionary, fine-grained descriptions, and global room style. The agent predicts material type, linear-RGB base color, roughness, metallic value, and specular strength, which are injected into the Blender script and bound to part constructors. Glass and mirror-like surfaces use shader overrides triggered by material type or part-name keywords; floors and walls use procedural shader nodes.

\paragraph{Stage 9 : Texture and decorative surfaces.}
For large or patterned surfaces, we use a high-capacity image generation model to synthesize texture maps for floors, walls, rugs, paintings, posters, and decorative panels, and inject them into the scene by augmenting the corresponding material node graphs with image-texture nodes. Planar decorative elements are assigned explicit UV mappings to ensure proper texture placement, while failed generations fall back to simplified prompts for more robust synthesis.

\paragraph{Stage 10: Lighting, rendering, and post-hoc correction.}
In the final stage, we complete the scene by setting up lighting, rendering parameters, and deterministic post-hoc corrections.
Conditioned on the input image and the generated scene, the VLM infers the overall lighting style, including the dominant illumination direction, window-driven natural light, possible artificial light sources, and ambient intensity.
These cues are then translated into Blender light objects and renderer settings to produce the raw executable scene $C_{\mathrm{raw}}$.
Before final rendering, we apply a deterministic correction pass that improves robustness without changing the semantic layout.
This pass fixes common implementation issues such as missing material assignments, invalid texture paths, unreasonable light intensities, incomplete camera coverage, and geometric artifacts.

For movable objects with boundary or overlap violations, we search for a nearby feasible placement:
\[
\mathbf{x}^{\star}_i
=
\arg\min_{\mathbf{x}\in\mathcal{N}(\hat{\mathbf{x}}_i)}
\|\mathbf{x}-\hat{\mathbf{x}}_i\|_2
\quad
\mathrm{s.t.} \;
B(o_i,\mathbf{x})\subseteq B_{\mathrm{room}},
\;
B(o_i,\mathbf{x})\cap B(o_j)=\emptyset .
\]
Here, $\hat{\mathbf{x}}_i$ is the generated position, $\mathcal{N}(\hat{\mathbf{x}}_i)$ is a local grid neighborhood, $B_{\mathrm{room}}$ is the room boundary, and the collision constraint is applied to nearby non-parent objects. In practice, this projection is implemented by deterministic local search with boundary clamping and stacking offsets for supported objects. The final program is 
$C=\mathrm{PostHoc}(C_{\mathrm{raw}})$,
where the final code $C$ serves as the complete representation of the generated 3D room scene, and can be directly executed in Blender to instantiate the full scene.

\section{Experiments}
\label{sec:results}

\begin{table*}[t]
\centering
\scriptsize
\setlength{\tabcolsep}{3.0pt}
\renewcommand{\arraystretch}{1.15}
\caption{
Our proposed benchmark evaluates different vision-language models for code-based 3D room generation from the top-down reference image, across various metrics, including visual content understanding, spatial reasoning, vision-to-code generation, and holistic scene quality.
}
\label{tab:vlm_benchmark}
\resizebox{\textwidth}{!}{%
\begin{tabular}{l cc ccccc cc ccc}
\toprule
\textbf{VLM}
& \multicolumn{2}{c}{\textbf{Visual Understanding}}
& \multicolumn{5}{c}{\textbf{Spatial Reasoning}}
& \multicolumn{2}{c}{\textbf{Code Generation}}
& \multicolumn{3}{c}{\textbf{Scene Quality}} \\
\cmidrule(lr){2-3}
\cmidrule(lr){4-8}
\cmidrule(lr){9-10}
\cmidrule(lr){11-13}
&
\makecell{Obj.\\Recall $\uparrow$}
&
\makecell{Func.\\Acc. $\uparrow$}
&
\makecell{Self\\Overlap $\downarrow$}
&
\makecell{Layout\\IoU $\uparrow$}
&
\makecell{Spatial\\ Relation $\uparrow$}
&
\makecell{Rotation\\Acc. $\uparrow$}
&
\makecell{Support\\Acc. $\uparrow$}
&
\makecell{Agent\\Completion $\uparrow$}
&
\makecell{Exec.\\Rate $\uparrow$}
&
\makecell{Image\\Similarity $\uparrow$}
&
\makecell{Scene\\Usability $\uparrow$}
&
\makecell{Aesthetic\\Quality $\uparrow$}\\
\midrule
Gemini3.1-pro~\cite{gemini31pro2026}      & 17.8\% & 15.3\%  & 8.4\%  & 16.8\% & 54.7\% & 78.0\% & 41.7\% & --     & 57.8\% & 2.49 & 3.24 & 2.51 \\
GPT-5.5~\cite{openai2026gpt55}          & 42.2\% & 71.7\%  & 14.5\% & 46.2\% & 50.8\% & 65.3\% & 52.6\% & --     & 42.2\% & 5.8 & 5.42 & 7.24 \\
\midrule
Gemini3-flash w/CaR~\cite{gemini31flashlite2026} & 58.9\% & 88.42\% & 2.57\% & 72.0\% & 76.9\% & 93.5\% & 93.5\% & 100\%  & 100\%  & 8.32 & 6.07 & 7.17 \\
Gemini3.1-pro w/CaR~\cite{gemini31pro2026}      & 55.5\% & 84.3\% & 3.3\% & 73.2\% & 79.8\% &93.6\% & 94.0\% & 100\%  & 95.5\% & 8.08 & 7.05 & 8.20 \\
GPT-5.5 w/CaR ~\cite{openai2026gpt55}             & 67.5\% & 72.54\% & 10.5\% & 66.7\% & 71.4\% & 92.2\% & 80.1\% & 71.1\% & 73.3\% & 7.28 & 6.00 & 7.52 \\
\bottomrule

\end{tabular}%
}
\end{table*}
\subsection{Experimental Setup}
We mainly conduct two evaluations.
First, we propose a benchmark to evaluate the effectiveness of our agentic harness with different VLMs, as well as the performance of different VLMs on this task.
Second, we compare our method with direct VLM generation and the agentic image-to-3D pipeline VIGA~\cite{viga}.
We further introduce image translation as re-rendering to demonstrate the future potential of this task, followed by a series of ablation studies.

\subsection{Benchmark for Top-down view image to 3D Room}
\paragraph{Benchmark Models}
\label{sec:benchmark_models}

We evaluate Code-as-Room (CaR) with three leading VLM backbones: Gemini-3 Flash~\cite{gemini31flashlite2026}, Gemini-3.1 Pro~\cite{gemini31pro2026}, and GPT-5.5~\cite{openai2026gpt55}.
These models are selected for their strong visual understanding, spatial reasoning, and code-generation abilities required for top-down image-to-3D scene synthesis.
We exclude Qwen-3.6~\cite{qwen3.6-27b} and Claude Opus-4.6~\cite{anthropic2026claude} from the main benchmark, as their preliminary results were often unusable for this task.
To assess the effect of our agentic workflow, we also test Gemini-3.1 Pro and GPT-5.5 under direct generation, where each model generates a complete Blender scene from the input image in a single call.

\paragraph{Benchmark Settings}
We evaluate our method on a test suite of 41 scenes covering diverse room types, scene complexities, and image styles.
The suite includes common residential spaces, such as bedrooms, kitchens, and living rooms, which are grouped into Simple, Middle, and Hard levels according to spatial scale and object density.
We further include specialized scenes, such as laboratories, barber shops, and cafes, to test robustness beyond standard home environments.
The input images span photorealistic photos, synthetic renderings, and abstract line drawings.
Since accurate ground truth is unavailable for such diverse inputs, we build annotations through a human-in-the-loop pipeline.
Specifically, Gemini 3.1 is first used within our agent workflow to generate coarse labels, which are then corrected by human annotators through a reverse code refinement tool that synchronizes visual edits with the underlying scene code.

\paragraph{Benchmark Metrics}
\label{sec:benchmark_metrics}

We evaluate different VLMs from four complementary aspects: visual understanding, spatial reasoning, code generation, and overall scene quality.
For visual understanding, we report object recall and functional accuracy, measuring whether the generated code recovers annotated objects and reconstructs major functional regions.
For spatial reasoning, we use self-overlap, layout IoU, spatial-relation consistency, rotation accuracy, and support accuracy to evaluate global layout alignment and local placement correctness.
For code generation, we report agent completion rate and Blender execution rate, measuring whether the full multi-stage pipeline can be completed and whether the final code runs successfully in Blender.
For overall scene quality, we use VLM-based scores on image similarity , scene usability, and aesthetic quality, together with human-study ratings on usability and error acceptability.

\begin{figure*}[t]
  \centering
  \includegraphics[width=\textwidth]{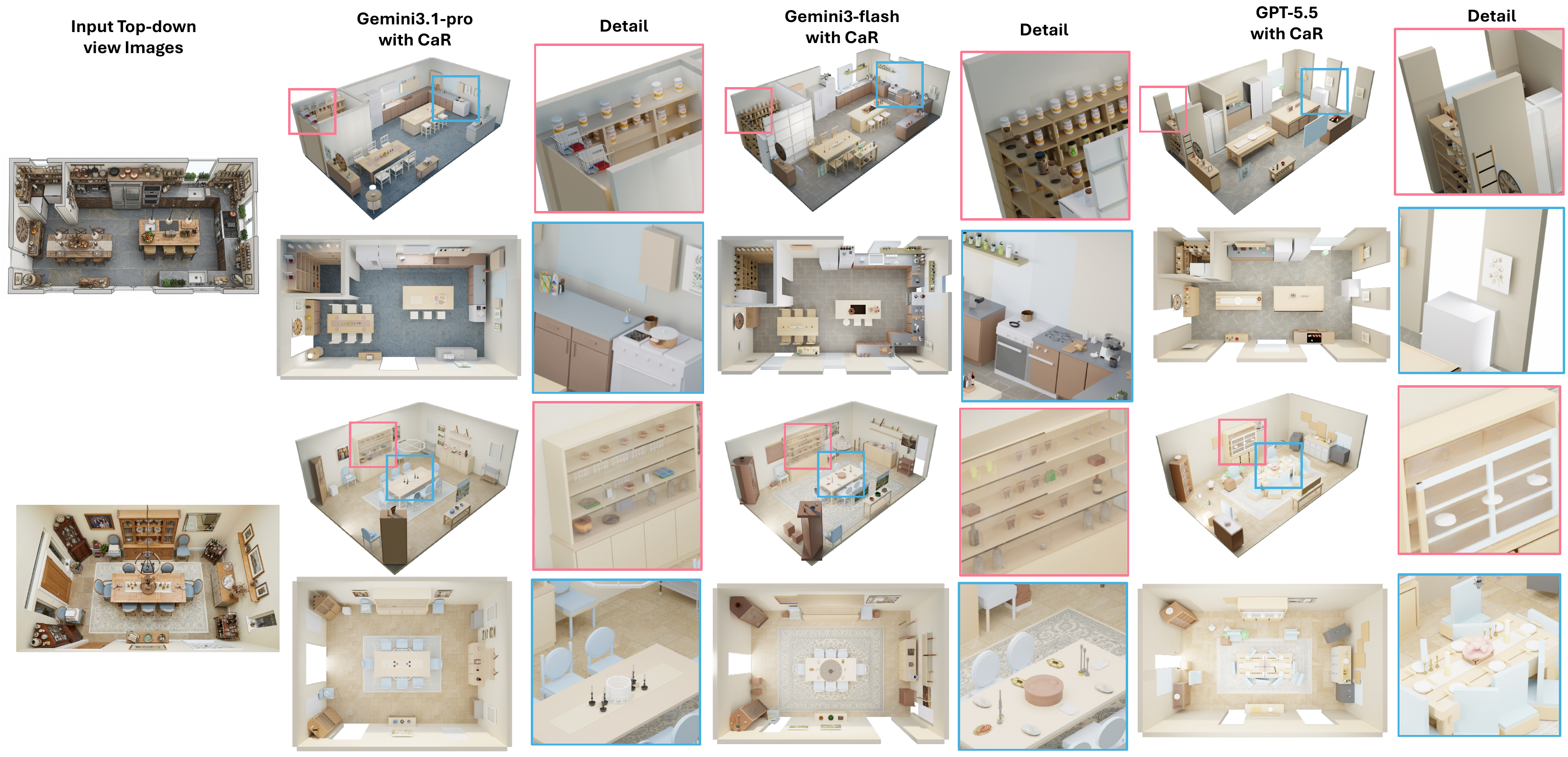}
\caption{%
  Qualitative comparisons corresponding to the benchmark results in Table~\ref{tab:vlm_benchmark}.
  With our harness in Code-as-Room, Gemini3-Flash, Gemini3.1-Pro, and GPT-5.5 can all generate executable 3D room code from top-down image inputs.
  The Gemini models achieve more stable results and demonstrate stronger spatial understanding than GPT-5.5, especially in object placement, layout consistency, and spatial relation preservation.
}
\label{fig:Benchmark_detail}
\end{figure*}

\paragraph{Benchmark Results}
Table~\ref{tab:vlm_benchmark} shows that our agentic design consistently improves VLM-based image-to-3D room generation.
Direct GPT-5.5 performs strongly in visual understanding and holistic quality, but still suffers from spatial inconsistency and unstable Blender execution.
Gemini models are weaker under direct prompting, yet become the most stable and competitive backbones when equipped with Code-as-Room.
These results indicate that staged decomposition, memory-guided reasoning, and visual feedback are critical for reliable 3D scene synthesis.

\begin{figure*}[t]
  \centering
  \includegraphics[width=1\textwidth]{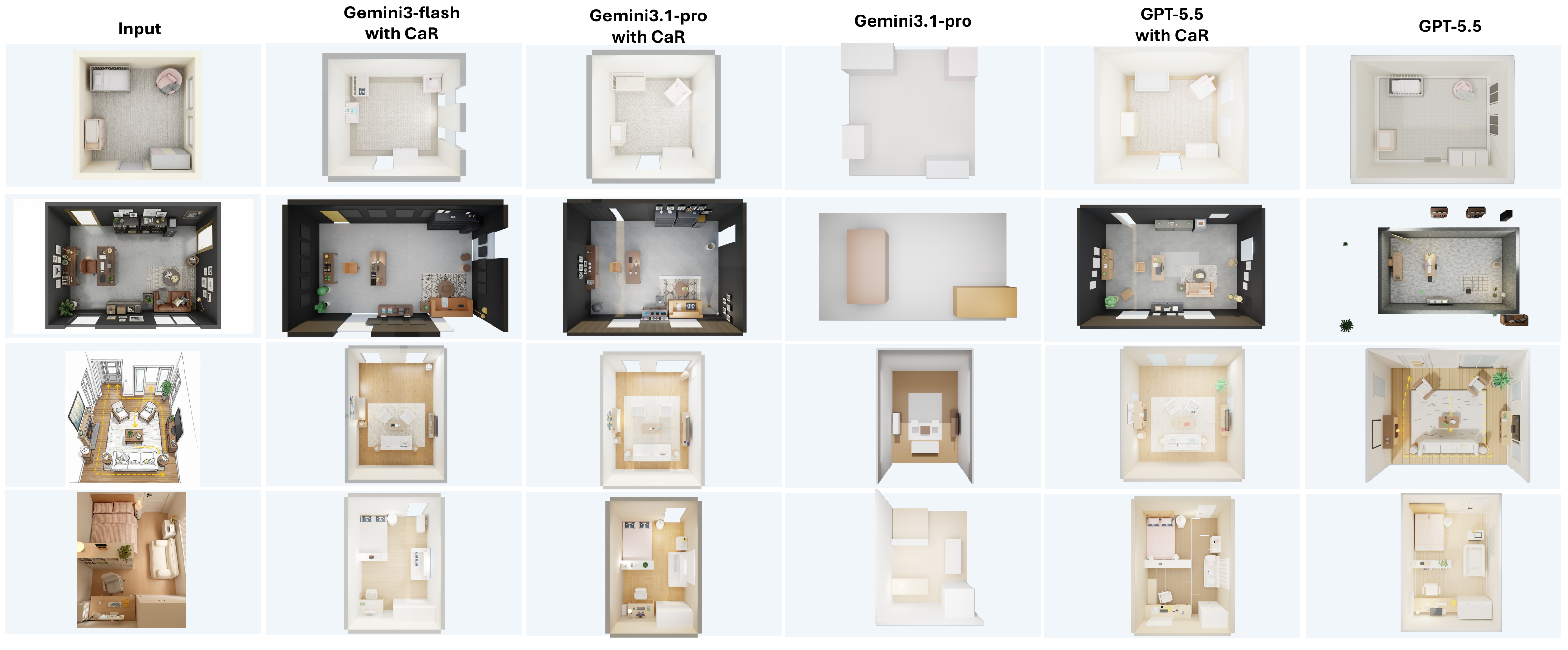}
  \caption{%
   Representative qualitative results corresponding to the benchmarks presented in Table~\ref{tab:vlm_benchmark}.
  }
\label{fig:Benchmark}
\end{figure*}

Figure~\ref{fig:Benchmark} further shows that direct VLM generation often produces incomplete or spatially unstable rooms, whereas Code-as-Room yields more complete structures, clearer functional regions, and better-aligned furniture layouts.
The detailed comparisons in Figure~\ref{fig:Benchmark_detail} show that Gemini-based variants recover richer local structures such as shelves, cabinets, decorations, and tabletop objects, while GPT-5.5 better preserves coarse layouts but tends to simplify small details.
Overall, successful image-to-3D room generation depends not only on the base VLM capability, but also on the structured agentic workflow used to organize it.

\begin{wraptable}{r}{7.5cm}
\centering
\caption{Human evaluation of overall scene quality.}
\label{tab:human_eval_results}
\resizebox{\linewidth}{!}{
\begin{tabular}{lcccc}
\toprule
\textbf{Method}
& \textbf{Sim.} $\uparrow$
& \textbf{Use.} $\uparrow$
& \textbf{Light} $\uparrow$
& \textbf{Accept.} $\uparrow$ \\
\midrule
\multicolumn{5}{l}{\textit{(a) Direct Generation Baselines}} \\
\quad Gemini3.1-Pro / Single-pass~\cite{gemini31pro2026} 
& 2.0 & 0.0 & 4.0 & 1.0 \\
\quad GPT-5.5 / Single-pass~\cite{openai2026gpt55} 
& 7.0 & 6.0 & 6.5 & 5.0 \\
\quad VIGA~\cite{viga} 
& 5.5 & 4.5 & 8.0 & 4.0 \\
\midrule
\multicolumn{5}{l}{\textit{(b) Code-as-Room Variants}} \\
\quad CaR w/ GPT-5.5~\cite{openai2026gpt55} 
& 7.5 & 7.0 & \textbf{8.0} & 6.5 \\
\quad CaR w/ Gemini3-Flash~\cite{gemini31flashlite2026} 
& 8.5 & \textbf{8.0} & \textbf{8.0} & \textbf{7.5} \\
\quad CaR w/ Gemini3.1-Pro~\cite{gemini31pro2026} 
& \textbf{9.0} & \textbf{8.0} & \textbf{8.0} & \textbf{7.5} \\
\bottomrule
\end{tabular}}
\end{wraptable}

\subsection{Human Evaluation}
\label{sec:human_eval}

We conduct a human evaluation with 20 experts, who rate each scene by similarity, usability, lighting alignment, and acceptability.
Acceptability measures whether the generated scene can be used after only minor manual corrections.
We compare direct VLM generation, Code-as-Room variants, and VIGA~\cite{viga}, with direct Gemini3.1-Pro and GPT-5.5 also serving as the Single-pass LLM baseline.

\begin{figure*}[t]
  \centering
  \includegraphics[width=1\textwidth]{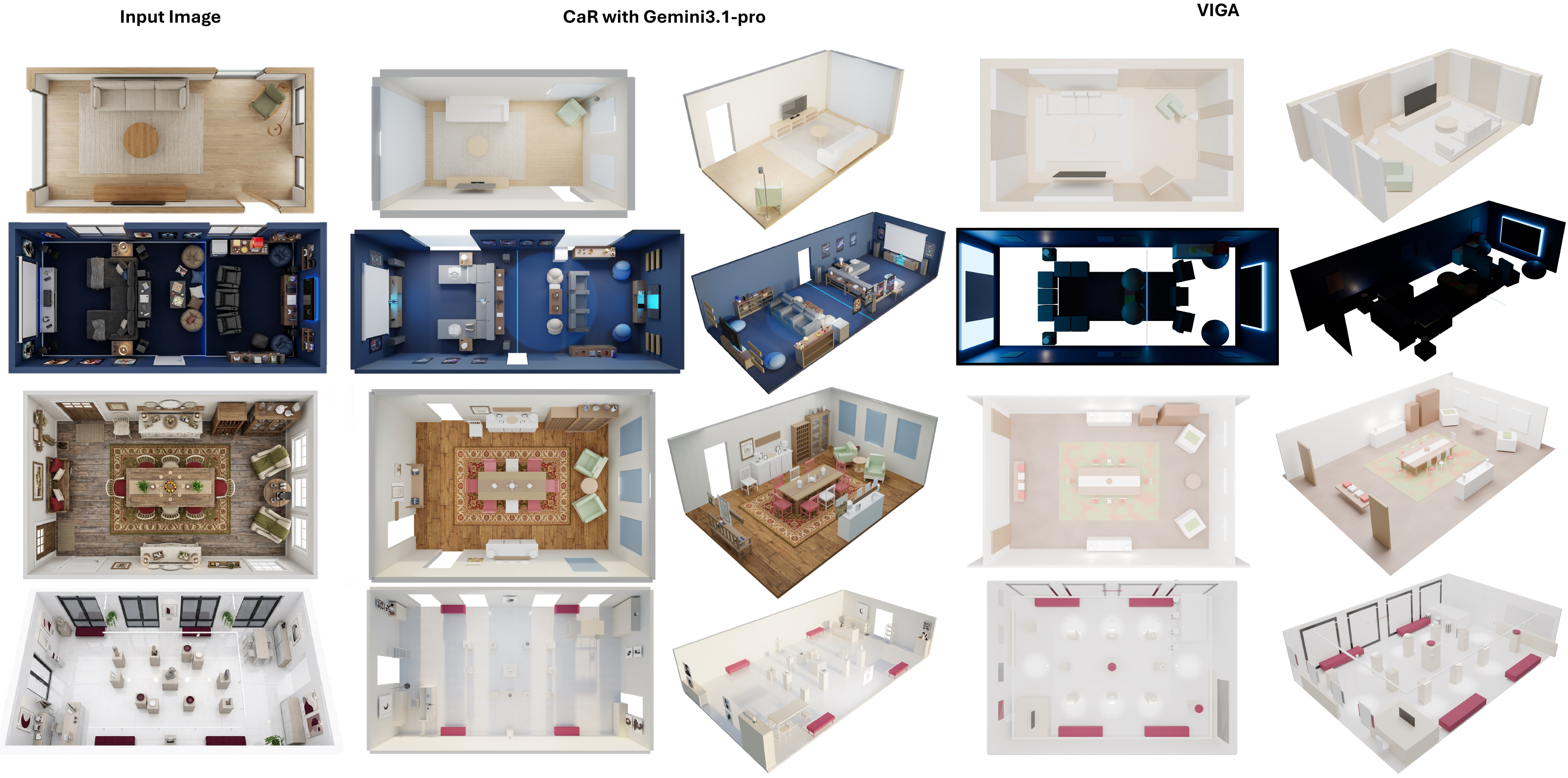}
  \caption{%
    Comparative performance analysis of Gemini 3.1-Pro integrated with CaR (ours) versus the VIGA framework.
  }
\label{fig:compare_gemini_viga}
\end{figure*}

Table~\ref{tab:human_eval_results} shows that Code-as-Room consistently improves human-perceived quality over direct generation and VIGA.
Our pipeline with Gemini3.1-Pro achieves the best similarity, usability, and acceptability scores.
Although VIGA obtains comparable lighting, its lower similarity and usability indicate weaker layout preservation and practical usability.
As shown in Figure~\ref{fig:compare_gemini_viga}, VIGA tends to generate template-like scenes with missing details and inaccurate object placement.
By contrast, Code-as-Room better preserves room proportions, furniture arrangements, and local semantics, demonstrating the benefit of our multi-stage agentic workflow.

\subsection{Scene Re-rendering}
\label{sec:rerender}

\begin{figure*}[t]
  \centering
  \includegraphics[width=1\textwidth]{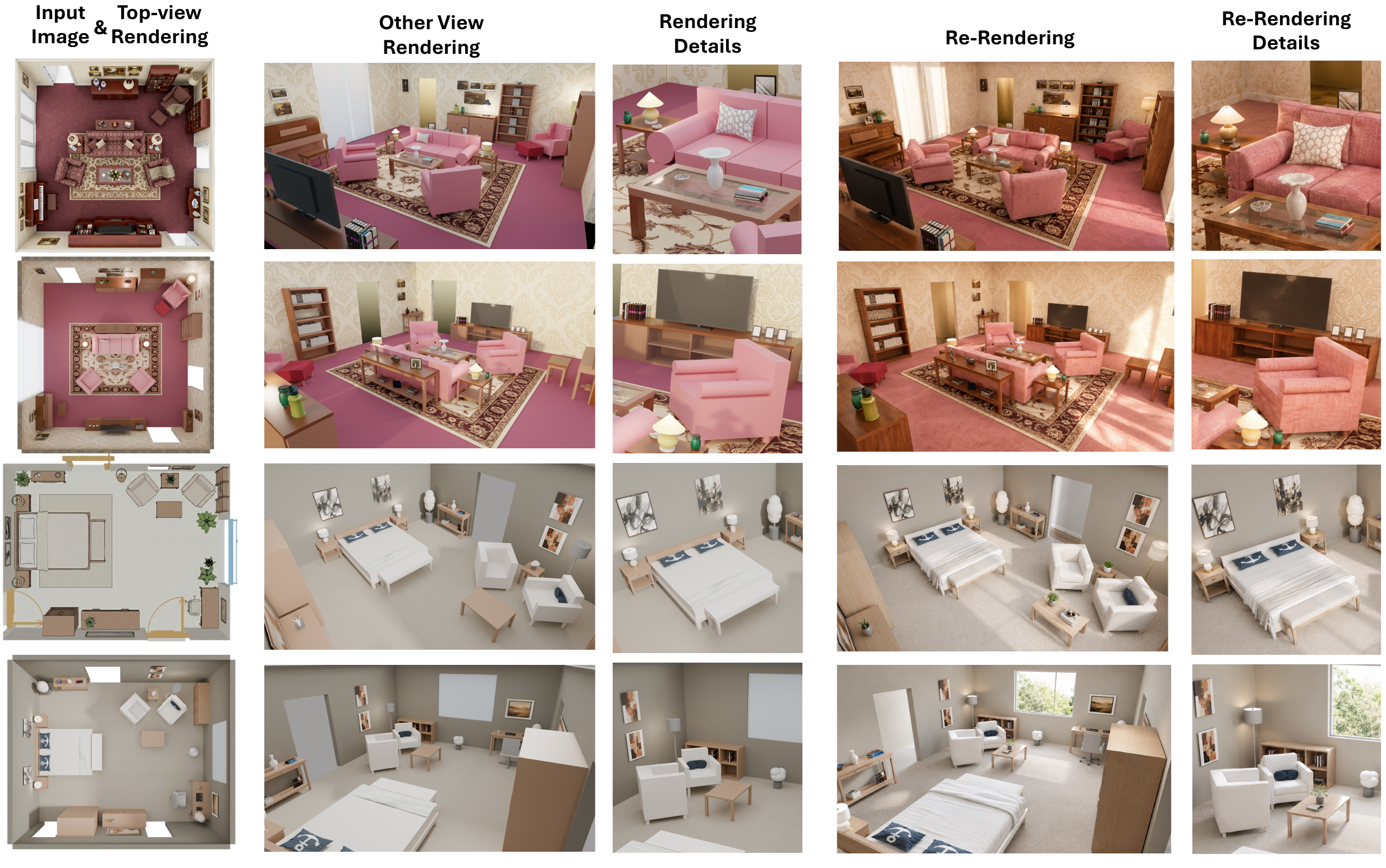}
  \caption{%
   Visual Enhancement Comparison: From Base 3D Scenes (left) to Realistic Re-rendering by GPT-5.5 (right).
  }
\label{fig:rerender}
\end{figure*}

To better show the potential of our generated results, we further apply image-level re-rendering to the Blender-rendered scenes.
While our 3D scenes are editable and geometrically consistent, many objects are still constructed from primitive shapes, which limits their visual realism.
Nevertheless, these scenes provide strong 3D priors, including room structure, object layout, spatial relations, and camera-consistent geometry.
We therefore use GPT-5.5~\cite{openai2026gpt55} to enhance the rendered images.
As shown in Figure~\ref{fig:rerender}, with stable geometric guidance from our 3D scenes, the re-rendering model produces more realistic materials, lighting, and object details without additional priors.
The enhanced results also preserve the original layout and maintain geometric and semantic consistency across multiple views, demonstrating that our generated 3D scenes can serve as effective structural priors for high-quality visual refinement.

\subsection{Ablation Studies}
\label{sec:ablation}

\begin{wraptable}{r}{7.5cm}
\centering

\caption{Ablation study on different components of Code-as-Room.}
\label{tab:ablation}
\resizebox{\linewidth}{!}{
\begin{tabular}{lccc}
\toprule
\textbf{Configuration}
& \textbf{Obj. Recall} $\uparrow$
& \textbf{Layout IoU} $\uparrow$
& \textbf{Rotation Acc.} $\uparrow$ \\
\midrule
\multicolumn{4}{l}{\textit{(a) Effect of Memory Mechanism}} \\
\quad w/o Memory
& 48.2\% & 58.0\% & 88.4\%  \\
\quad Full Model (Ours)
& \textbf{55.5\%} & \textbf{73.2\%} & \textbf{93.6\%}  \\
\midrule
\multicolumn{4}{l}{\textit{(b) Effect of Visual Feedback Iterations}} \\
\quad w/o Visual Feedback (0 iter.) & 33.8\% & 64.0\% & 71.9\% \\
\quad Feedback $\times$3 & 35.6\%& 65.7\% & 73.2\%  \\
\quad Feedback $\times$5 (Ours) & 38.4\% & \textbf{66.2\%} & \textbf{75.4\%}  \\
\quad Feedback $\times$10 & \textbf{39.1}\% & 64.2\% & 72.6\%  \\
\bottomrule
\end{tabular}}

\end{wraptable}

\begin{figure*}[t]
  \centering
  \includegraphics[width=1\textwidth]{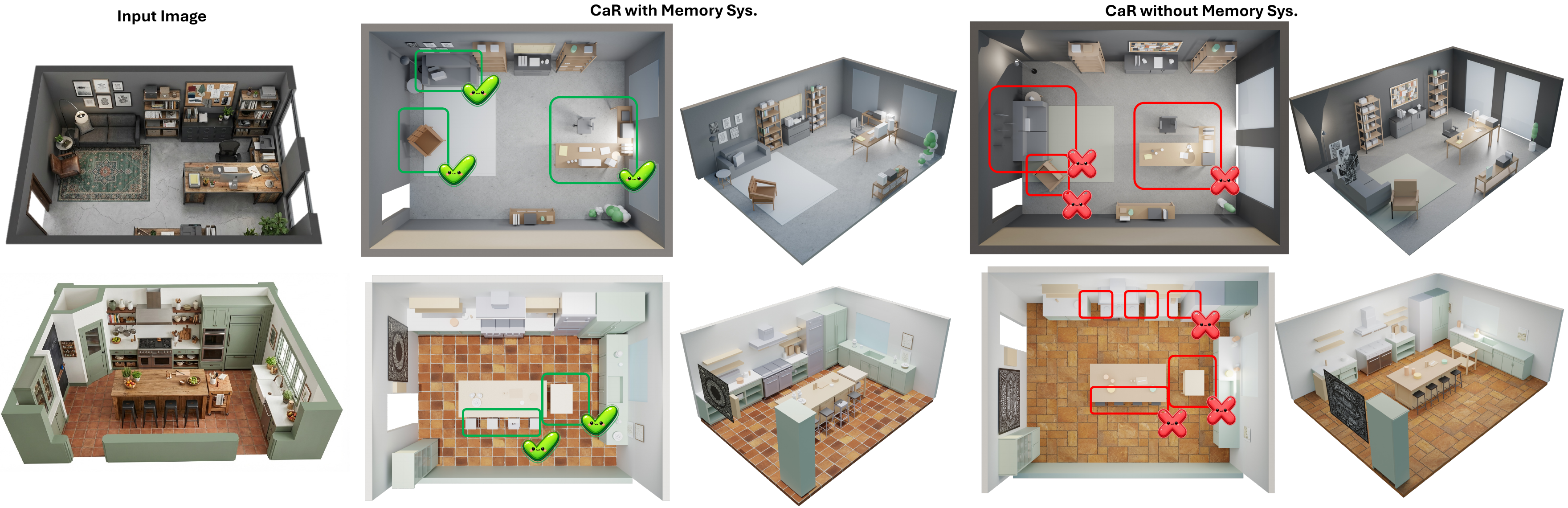}
  \includegraphics[width=1\textwidth]{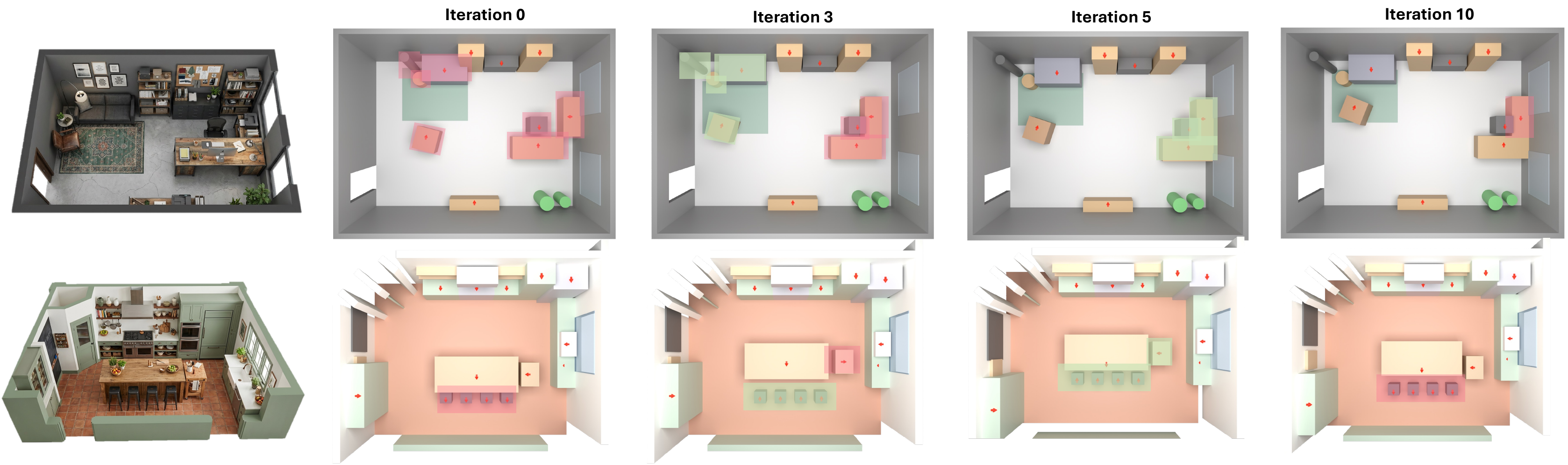}
  \caption{%
    Ablation study results evaluating the impact of the memory system and visual feedback iterations (referencing data in Table~\ref{tab:ablation}).
  }
\label{fig:ablation_memory_system}
\end{figure*}

We conduct ablation studies to analyze the memory mechanism and the visual feedback loop in Code-as-Room.
All variants use the same VLM backbone and input images for fair comparison.
As shown in Table~\ref{tab:ablation}, removing memory degrades all metrics.
Without cross-stage memory, later stages cannot reliably reuse earlier image-derived information, leading to missing objects and weaker layout preservation.
The drop is especially large in Layout IoU, showing that memory is important for maintaining spatial consistency across stages.
For visual feedback, performance improves from 0 to 5 iterations, indicating that intermediate renderings help correct object omissions, placement errors, and rotation misalignment.
However, increasing the feedback rounds to 10 reduces Layout IoU and rotation accuracy, suggesting that excessive revision may introduce layout drift or over-correction.
Therefore, we use five feedback iterations as a balance between generation quality and refinement cost.
Figure~\ref{fig:ablation_memory_system} provides qualitative examples of these effects.

\section{Conclusion}
\label{sec:conclusion}

In this paper, we presented \textbf{Code-as-Room}, an MLLM-based agentic framework for synthesizing realistic and functional 3D indoor rooms from top-down reference images. Unlike text-driven room generation methods, our framework uses the input image as an explicit spatial prior and represents the generated scene as executable Blender code, enabling editable, renderable, and structured 3D room synthesis. To improve the stability of holistic room generation, Code-as-Room decomposes the task into a principled multi-stage pipeline that progressively performs visual parsing, spatial reasoning, scene layout construction, object-level code generation, and material-lighting refinement. A cross-stage memory module is further introduced to preserve scene information across stages and reduce context forgetting in long agentic workflows. We also built a dedicated benchmark for code-based 3D room synthesis, covering visual understanding, spatial reasoning, code generation, and holistic scene quality. Comprehensive experiments demonstrate that the proposed execution harness substantially improves different MLLM backbones and outperforms existing agent-based baselines in generating coherent, usable, and visually aligned indoor scenes. 
\paragraph{Limitations and future work.}
\label{sec:limitations}
First, it currently targets global 3D scene synthesis from top-down view images and is not yet optimized for arbitrary-view inputs, which limits its applicability to more general real-world settings.
Second, many real-world objects remain difficult to faithfully generate with procedural code, due to the limited alignment between code-generation models and 3D asset.
Thus, retrieval-based asset insertion may still be needed for higher geometric fidelity.
Second, although image re-rendering can improve visual realism while preserving multi-view geometry and semantics, current video models still struggle with high-quality and temporally consistent re-rendering, especially for trajectories longer than five seconds.
In future work, we will first explore video generation models as neural renderers for Code-as-Room to produce more realistic and coherent scene visualizations.

\begin{figure*}[t]
  \centering
  \includegraphics[width=1\textwidth]{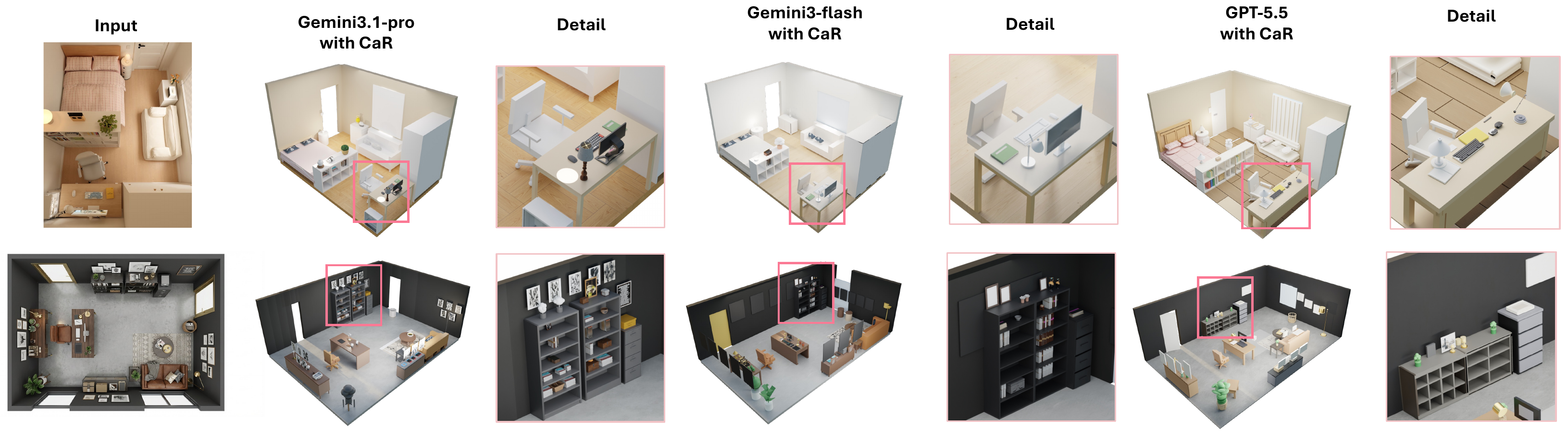}
  \caption{%
   Detailed qualitative analysis and performance comparisons relative to the benchmarks in Table~\ref{tab:vlm_benchmark}.
  }
\label{fig:Benchmark_detail}
\end{figure*}

\clearpage

{
    \small
    \bibliographystyle{plainnat}
    \bibliography{neurips_2025}
}


\newpage

\end{document}